# Large Language Models Have Not Yet Transformed Science

**Ron Bekkerman** **Nichole Li** **Bahareh Kargar** **Mark Somers**

*New Jersey Institute of Technology, Newark, NJ, USA*

*{ron.bekkerman,nichole.li,bahareh.kargar,mark.somers}@njit.edu*

## Abstract

The emergence of large language models (LLMs) in late 2022 has raised questions about the future of science, but assessing their impact requires understanding how scientific practices were already evolving. We analyze 2.5 million full-text papers across five disciplines from 2015 to 2025, tracing trends over a broad spectrum of lexical, structural, epistemological, and social dimensions. Aside from frequency shifts in some words, most pre-2023 trends continued with limited detectable breaks. Among the notable exceptions, growth in computer science coauthor teams accelerated significantly in Asia, while apparent declines in biomedical topic diversity largely reflected standardization in article formats. Our study establishes an empirical foundation for monitoring the health of global science as LLM adoption deepens.

## Introduction

Large Language Models (LLMs) are used widely in scholarly research, with tangible benefits to scientists including increased productivity and visibility [1]. Although LLMs appear to enhance the work of individual scientists, questions have been raised about their influence on science as a whole. Indeed, Hao et al. [2] raise the prospect of an AI paradox in which researchers' productivity gains come at the expense of research quality. The appearance of AI slop in scientific papers [3,4] has been linked to diminished rigor and accuracy of research [5]. Thus, LLMs' effects on scientific language are hypothesized to spill over to epistemology fostering a scientific monoculture [6] that limits novelty and innovation [7].

Gaining a better understanding of the AI paradox involves quantifying its scale and scope. Language is central to scientific communication so considerable effort has been directed toward measuring the effect of LLMs on scientific language including paper vocabulary [8], the language quality [1,9], readability [10], the quality of thought and problem definition [11,12], and novelty and creativity [7]. Attention has also been drawn to the effect of LLMs on research collaboration and productivity in non-English speaking countries [5, 13-16].

The broad scope of potential macro-level shifts following the public emergence of LLMs requires a comprehensive analytical framework. We developed and applied such a framework, aligned with the scale and multi-dimensional nature of LLMs' footprint on scientific research as reflected in scientific writing (Figure 1). It extends work in this domain by providing new insights into whether scientific writing exhibits systemic degradation post-2022 despite the tangible benefits of LLMs for individual scholars - that is, we address the broad implications of the AI paradox. As indicated in Figure 1, language is central to our framework. The effect of LLM emergence on language was assessed across nested attributes starting with the narrowest view on AI-favored keywords, and then broadening the scope to lexical diversity [17,18] and density of scientific terminology. Our framework allows us to examine whether LLM usage affects lexical diversity and replaces scientific terms with AI-favored surrogates [5,8,17,19].

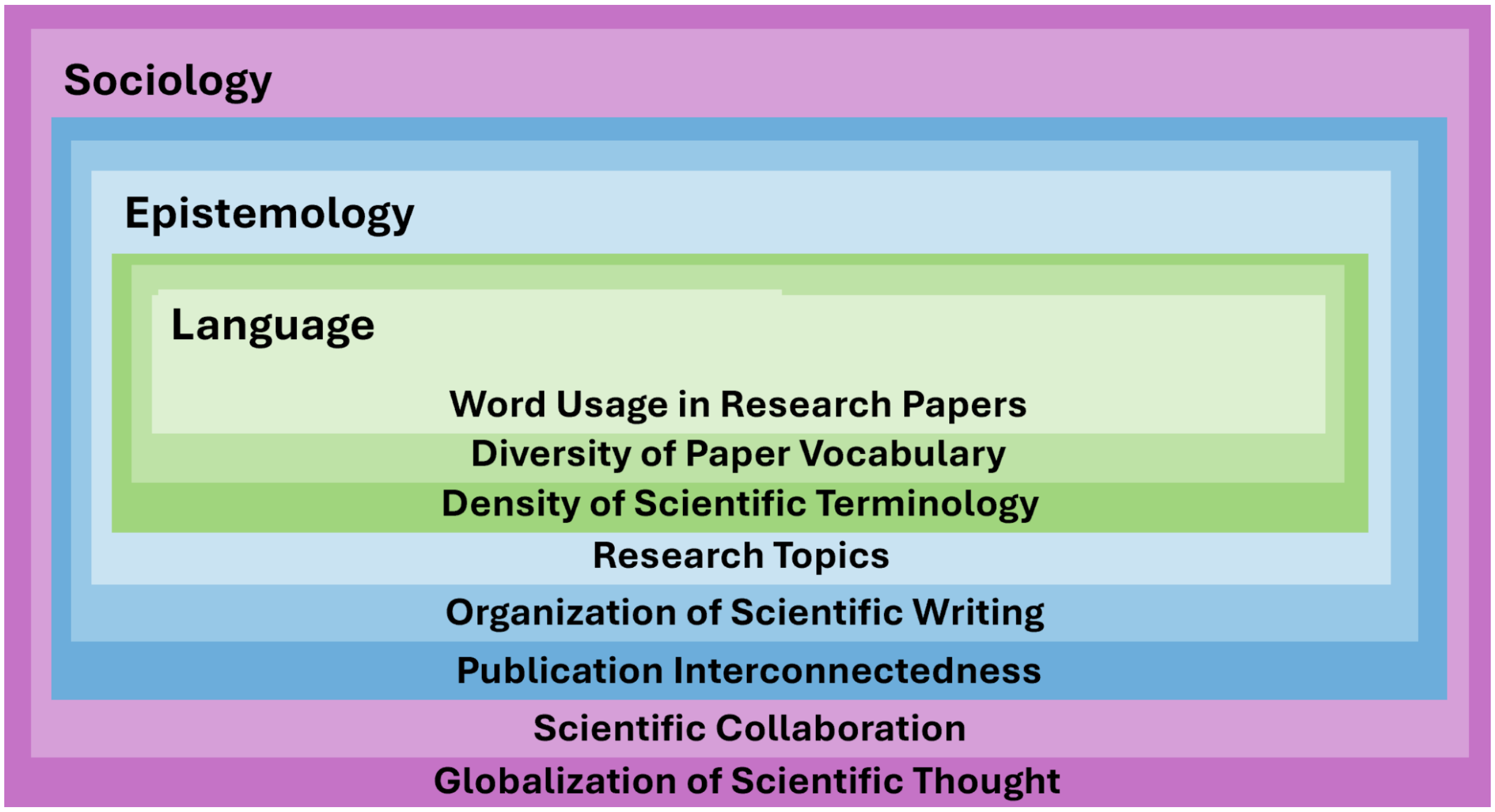

**Figure 1: A comprehensive framework for analyzing changes in scientific research following the emergence of LLMs, as reflected in scientific writing.** The framework spans the broad spectrum of relevant attributes, organized into nested lexical, epistemological, and social dimensions. We measure a limited number of representative attributes within each dimension, from the narrowest (word usage) to the widest (global landscape of science). The comprehensiveness of our framework lies in the breadth of dimensions covered, while the selected attributes provide an initial operationalization of this broad spectrum and a foundation for further work.

Next, the connection between language and epistemology was assessed by analyzing the extent to which research is becoming homogenized within disciplines through reduced topic diversity, as well as post-2022 structural shifts in the organization of scientific papers (which is tied to scientific thought) including paper interconnectedness. Finally, we checked whether the public introduction of LLMs catalyzed the expansion of opportunities for collaboration and participation from non-English-speaking regions, facilitating the democratization of scientific research.

Our study was conducted on a large publicly available scientific repository: about 2.5 million full text papers published from 2015 to 2025 in the following disciplines: Computer Science (CS), Mathematics, Physics, Health Sciences and Life Sciences. About fifty percent of them (Health and Life Sciences) were collected from the PubMed Central (PMC) database, while the remaining papers were collected from the arXiv database (Computer Science, Math and Physics). Together, these disciplines span theoretical, experimental, and applied sciences, providing a broad view of global scientific activity and enabling us to examine changes across multiple types of research practices. Our decade-long measurement window allows a large-scale analysis of research trends without getting unnecessarily deep into the history of science.

The AI paradox is grounded in the assumption that LLMs are changing scientific culture. Although LLMs are a disruptive technology with epistemological implications, it has been unclear whether they are at the center of a nascent or emerging paradigm shift that redefines how scientific research is designed, conducted, and communicated. It has also been unclear which side prevails in the tension between the LLMs and current practices. We do know that paradigms change slowly unless there is an external shock [20]. LLMs can be considered a shock but whether the shock is strong enough to disrupt scientific research remains to be seen. Our comprehensive framework is designed to help answer these questions.

Our results indicate that the systemic impact of LLM emergence on scientific research remains limited. Many established trends for multiple attributes of language, epistemology, and research collaboration across multiple disciplines were not altered post-2022. So far, the trajectory of scientific research follows established trends guided by prevailing paradigms. Thus, our study provides evidence consistent with the resilience of scientific culture in the face of challenges to accepted practices from the side of LLMs. We believe that future studies using comprehensive frameworks such as ours are important for monitoring the systemic footprint of LLMs on scientific research over time.

## LLMs Affect Select Word Usage, Not the Entire Vocabulary

Recent studies have analyzed characteristic vocabulary shifts in post-LLM literature by using predefined or manually curated word lists, typically applied to paper abstracts [8,9,21]. Our study confirms and extends these findings. Instead of starting with a list of preselected words, we programmatically identify words whose usage was likely affected by the introduction of LLMs, and test their usage dynamics on our dataset of 2.5 million full paper texts.

The goal is to identify words whose frequencies shift systematically following the introduction of LLMs, in ways that distinguish their pre-LLM and post-LLM usage. Starting with 20,000 most common English words that cover practically the entire everyday English vocabulary, we apply a domain-aware lemmatization technique that yields a list of 14,484 lemmas. For each lemma, we fit segmented linear regression to its annual per-million-word frequency series with a breakpoint at 2022 [22].

Of the 14,484 lemmas analyzed, 796 show a significant post-2022 surge, 71 show a rise-and-drop pattern, and 199 show a significant decline. Surge words (such as *establish* and *underscore*) are sophisticated, analytical terms that LLMs tend to prefer. Decline words (such as *good* and *happen*) are short, plain alternatives that LLM-edited prose often avoids. The trajectory of words like *delve*, widely noted as a marker of GPT-style writing [8], illustrates the rise-and-drop pattern: nearly absent before 2023, they peaked in 2024 and sharply declined in 2025, consistent with both model updating and growing public awareness of AI slop.

Figure 2 presents these trends by discipline. The signal is strongest in Computer Science (blue), with the largest post-2022 changes across all three word types, and the rise-and-drop pattern being especially profound. Health Sciences (red) and Life Sciences (purple) show comparable shifts from higher baselines; Physics (green) follows at a smaller magnitude, and Mathematics (orange) changes least. This ordering is consistent with known patterns of LLM adoption [23]: Computer Science researchers, as both developers and early adopters of LLM tools, leave the strongest imprint, while mathematical writing, constrained by formal notation, shows practically no effect.

The broad scope of this analysis provides evidence of systematic vocabulary shifts following the introduction of LLMs, with AI-favored keywords waxing and waning alongside ongoing model development. These vocabulary shifts are not confined only to abstracts but are also evident throughout the main text. Surge words are approximately twice as frequently used in abstracts as in paper bodies, but this ratio is stable before and after the LLM introduction, suggesting that the ratio reflects the summary- and emphasis-oriented nature of abstracts rather than an LLM-specific concentration effect (see Supplementary Material). Nevertheless, the size of the affected vocabulary (slightly over a thousand words) is negligibly small in comparison to the entire vocabulary used in our dataset (dozens of millions of unique words), which leaves the overall lexical composition of scientific writing largely unchanged.

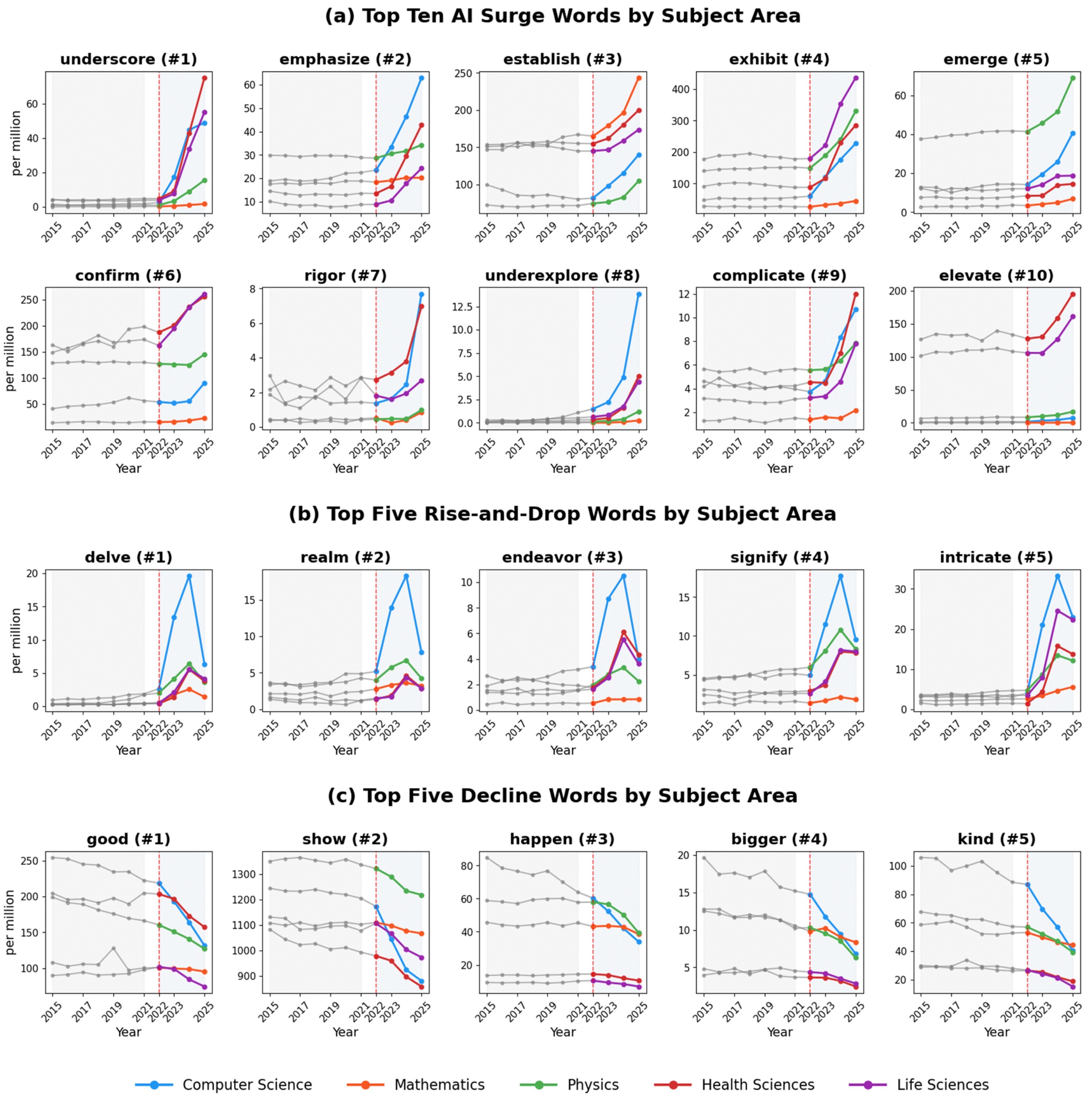

**Figure 2: LLM-affected words, by subject area.** Annual per-million-word frequency, by subject area, for the top ten Surge words *(a)*, top five Rise-and-Drop words *(b)*, and top five Decline words *(c)*. Those words are top-ranked in the pooled-corpus analysis; no separate subject-level ranking was performed. Words within each panel are ordered by pooled fold change (#1 = strongest overall signal).

## Scientific Language Appears More Scientific, but Not Necessarily More Diverse

The prevalence of AI-favored keywords and language has been associated with increased lexical diversity and decreased scientific rigor in scientific papers. LLMs have been characterized as generating language that appears plausible, but lacks the precision and rigor of scientific research [5]. To address this issue, we analyzed trends in lexical diversity and the usage of scientific terminology from 2015 to 2025.

Lexical diversity is traditionally measured by Type-Token Ratio (TTR) which is defined as the ratio of unique words in a document (i.e. the vocabulary size) to the total number of word occurrences (the document length) [24]. TTR suffers from a length-dependency bias: its values are inherently lower for longer documents, which makes direct comparisons across documents of varying lengths methodologically problematic. To overcome this problem, we use Moving Average TTR (MATTR) [25] which is a mean TTR across sequential moving windows of a fixed length (e.g. 50 words [17]). We present MATTR at 50 in Figure 3 (first panel). As we show in Supplementary Material, there is no statistically significant slope change in 2022 for any discipline. Still, Computer Science behaves surprisingly differently from the other four disciplines: its lexical diversity grows rapidly since 2015, then slows down around 2020-2022, and then restarts its growth in 2023.

To provide some additional context, we plotted vocabulary sizes and paper lengths (in number of words) – second and third panel of Figure 3, respectively. Paper vocabularies show an uninterrupted growth across the five disciplines, while paper lengths grow particularly steeply in 2020-2021, during the COVID-19 pandemic. Computer Science researchers wrote much longer, but not lexically richer papers during COVID, and then appear to return to the pre-COVID pattern (stable paper length, lexical diversity growth).

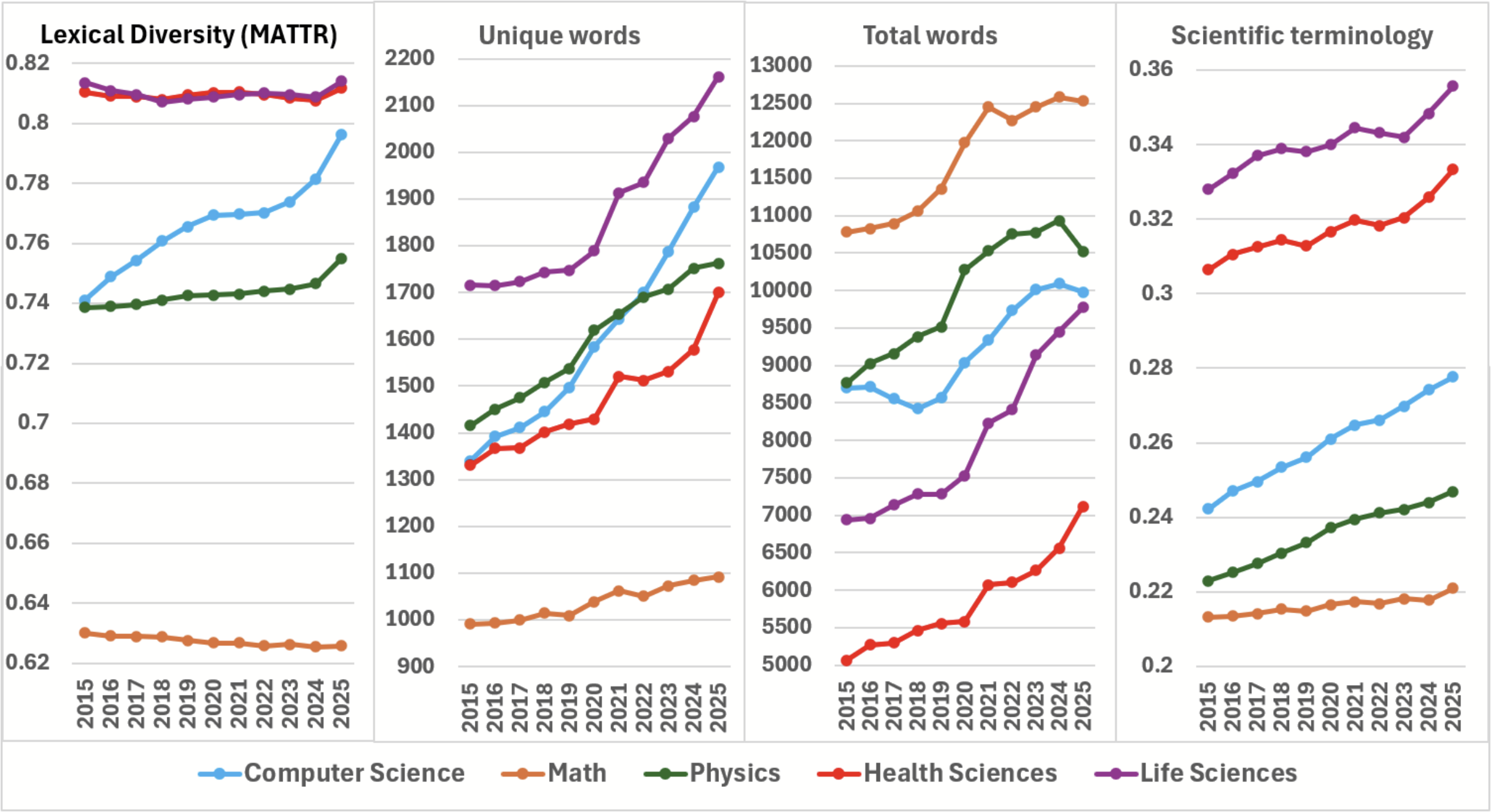

**Figure 3: Measuring the dynamics of scientific vocabulary.** (1) average lexical diversity, measured as MATTR with the sliding window of 50 words; (2) average vocabulary size per paper, measured as the number of unique words, excluding math notation; (3) average paper length, measured as the total number of words, excluding math notation; (4) average density of scientific terminology, measured as the ratio of the number of scientific terms in the paper to its vocabulary size.

The density of scientific terminology in research papers is assessed by computing the ratio of the number of scientific terms to the number of unique words (Figure 3, right). Smaller ratios are indicative of less use of scientific terminology. We utilize NJIT's PaperBasket [26] infrastructure to extract scientific terminology from paper texts (see Supplementary Material). We present clear evidence of a consistent upward trend in scientific terminology usage in all disciplines. Thus, LLMs did not appear to "dilute" scientific papers with AI slop. Rather, scientific papers appear more scientific over time.

These results, coupled with our findings for AI-favored keywords, provide evidence that the emergence of LLMs has not inflated lexical diversity or diminished research communication quality with less rigorous language. The macro-level impact of AI-favored keywords on overall scientific language remains strictly constrained.

## Science Stays Topically Diverse, with One Apparent Exception

The connection between language and epistemology is well established [27]. Concerns have been raised that LLMs are creating a scientific monoculture which limits novelty and innovation [7]. At issue is whether LLMs are framing how research problems are identified and investigated, resulting in a limited number of topics and avenues of inquiry [28]. We analyzed topic diversity dynamics, before and after the introduction of LLMs, by measuring Jensen-Shannon (JS) distances between scientific papers [29]. For each year between 2015 and 2025, we measured the mean JS distance between pairs of papers, as well as the standard error of the mean (see Supplementary Material). Two analyses were conducted: papers were represented as probability distributions over their words and over scientific terminology extracted from their texts. The JS distances are bounded between 0 (full overlap in words or terminology) and 1 (no overlap). In practice, JS distances are close to 1 as two randomly selected papers hardly overlap.

Mean pairwise JS distances for word and terminology distributions are presented in Figure 4. Graphs are practically flat for Physics and Math, indicating their topic diversity did not change post-2022. The decline in Computer Science is remarkable - and almost linear. We analyzed the most commonly used terminology in CS papers and concluded that the increasing similarity of topics is a consequence of established patterns in Computer Science research, as the discipline coalesces around emerging areas such as deep learning and generative AI. Apparently, this trend was initiated way before the introduction of LLMs in 2022 - Computer Science is the discipline that introduced LLMs - and appears to have slightly slowed down since 2021.

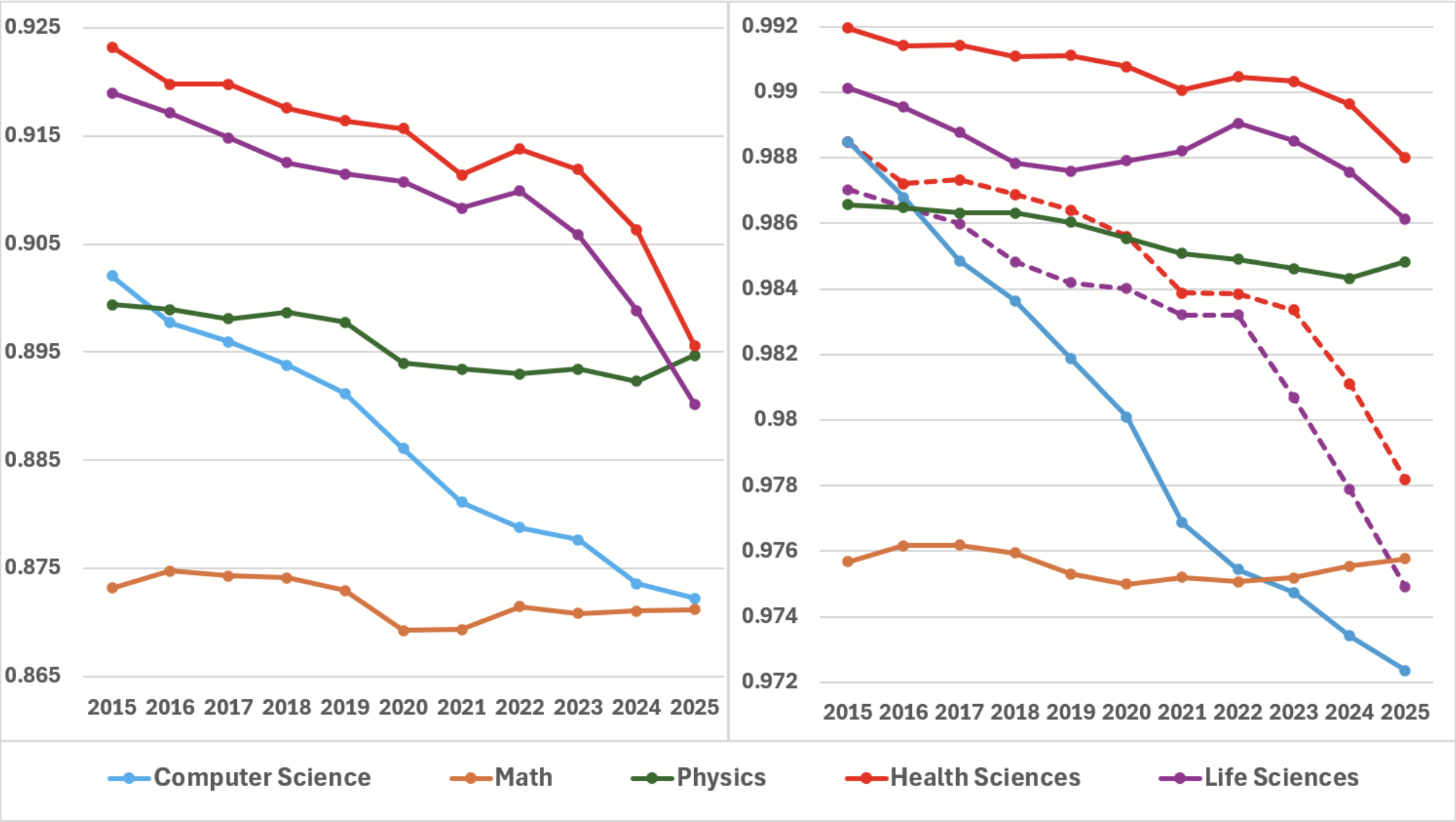

**Figure 4: Topic diversity (in terms of mean pairwise JS distances).** (Left) papers represented as word distributions; (right) papers represented as terminology distributions. For Health and Life Sciences, two versions are shown. The dotted lines are topic diversity results obtained over original, unfiltered terminology distributions; the solid lines are the results after 105 boilerplate keyphrases were filtered out. Note that the standard errors of the mean are very small: if they were shown as error bars, the bars would have been covered by the tick marks.

Health and Life Sciences showed a decline in topic diversity, which intensified substantially post-2022, coinciding with the introduction of LLMs. To understand this phenomenon, we analyzed the most common terminology used in the PMC literature. A large portion of papers' similarity does not come from their topics, but rather from their format. Many biomedical papers contain standard boilerplates that cover publishing metadata and licensing, open source disclosures, and contributor roles. In recent years, there was a movement for paper format unification in biomedical literature, which made more papers artificially similar to each other.

Filtering the top 105 boilerplate terms substantially mitigated these structural artifacts, bringing the topic diversity baselines for Health and Life Sciences closer to the steady trajectories, similar to the ones of Math and Physics. However, because keyphrase filtering cannot exhaustively capture every publisher-specific formatting update, minor residual noise from ongoing metadata standardization is likely to persist. Note that we adjusted only terminology distributions (Figure 4, right) but not word distributions (Figure 4, left) so the word distribution graphs still reflect the artifacts of paper formatting.

Formal tests of detected trends, including equivalence-type and Bayesian analyses and their sensitivity to key choices, are reported in the Supplementary Material.

## Trends of Paper Structure Persist Post-2022, but Interconnectedness Inflects

Language also influences epistemology by shaping how thoughts are organized [28]. The impact of LLM emergence on the structural organization of scientific papers is an important, under-researched topic. It provides key insights into whether research communication is undergoing structural standardization in the LLM era. As research communication is discipline-specific, we conducted our first analysis on Computer Science (before generalizing to all five disciplines). Our analysis includes section headings, section order, and section lengths. The analysis covers 437,980 CS papers published on arXiv between 2015 and 2025.

To define the structure of research papers, we use nine section titles representing the core structure of a typical Computer Science paper: “Abstract”, “Introduction”, “Related Work”, “Method”, “Experiments”, ”Results”, ”Discussion”, “Conclusion”, and “References”. Ancillary sections, such as ”Acknowledgements”, “Appendix”, “Datasets”, and “Limitations” are excluded. Each section is associated with a dictionary of common naming variants found in the corpus. For example, “Methodology”, “Methods”, and “Proposed Method” all map to “Method”. The dictionary contains over 80 variants covering the nine standard section titles. For each paper, we match section titles against this dictionary to determine which sections are present, their order of appearance, and their lengths in words. Section lengths are measured as the word count from their title to the next detected section title. For section ordering, we compute the percentage of papers defined by a 5-section template (Abstract, Introduction, Related Work, Conclusion, References) and a 7-section template (Abstract, Introduction, Related Work, Method, Experiments, Conclusion, References). For length variability, we report the mean and standard deviation of section word counts per year.

Our analyses indicate that changes in the structure of papers are present, but that they follow a well-defined trend that was not disrupted by the introduction of LLMs. As indicated in Figure 5(a), the volume of papers with only 1 to 3 sections dropped from 17% to 4%, while the segment of papers with 8 or more sections grew from 5% to 30%. The sections with the strongest growth were “Related Work” (32% to 66%), “Method” (11% to 45%), and “Experiments” (31% to 59%). These findings indicate that Computer Science papers are adopting a common set of standard section names following established trends within the discipline.

The trend toward convergence in the structure of research papers is affirmed by analyses of section length and section ordering. With respect to section order, the 5-section template grew from 25% to 55% while the 7-section template grew from 1% to 19% (Figure 5b). Changes in section lengths trade off against the increase in the number of sections (Figure 5c). “Introduction” and “Related Work” – sections that are traditionally considered “soft” and less technical – shortened significantly from 2141 to 1236 words and from 1955 to 1267 words, respectively. In contrast, technical sections “Method” and “Experiments” remained relatively stable. This trend corresponds to the earlier established fact that papers look more scientific. Interestingly, “References” grew dramatically from 682 to 1261 words. Analysis of standard deviations of section lengths (Figure 5d) showed a trend toward convergence, with the exception of “References”.

The exceptional dynamics of “References” in Computer Science led us to investigate this phenomenon globally, by determining the annual average number of references across the five disciplines. To complete the picture, we extended this analysis to the average number of tables and figures (Figure 5, bottom row). The average number of references grows across the board, while the growth seems to plateau in Computer Science and Physics and to accelerate in Health and Life Sciences.

CS and Physics researchers, being early adopters of AI technologies, were likely to use AI for literature reviews prior to the LLM introduction, which led to an increase in the number of references. In recent years, however, the trend toward paper structure convergence led to a unification in the number of references. Biomedical researchers appear to be late adopters of AI technologies, who may have more recently started utilizing LLMs for literature review leading to an increase in reference counts. Thus, AI technologies in general, and LLMs in particular improve research interconnectedness.

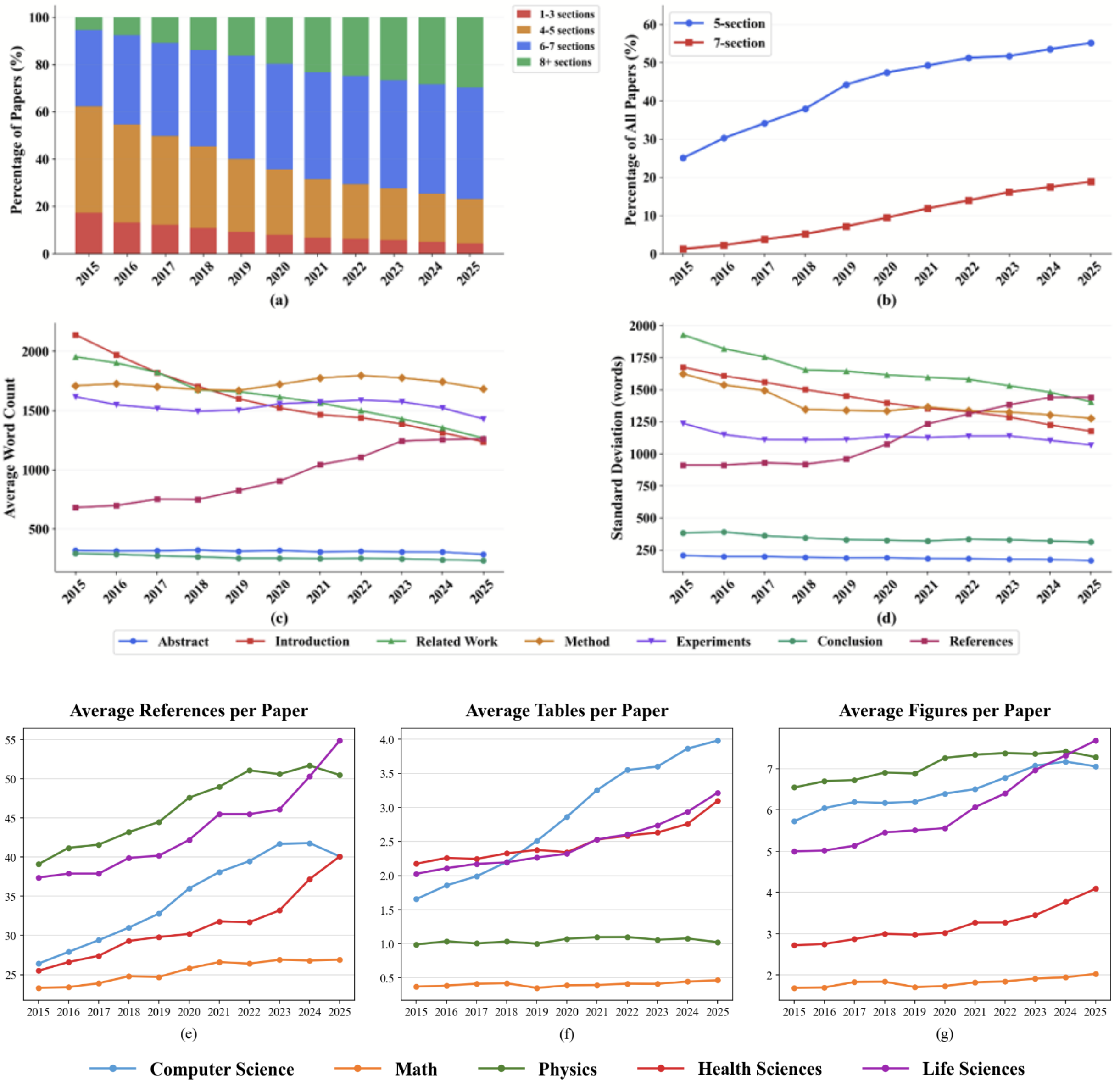

**Figure 5: Measuring structure of scientific writing.** (a) Distribution of the number of sections per paper; (b) Adoption of order consistency for the 5-section and 7-section templates; (c) average section lengths; (d) standard deviation of section lengths; (e) average number of references per paper; (f) average number of tables per paper; (g) average number of figures per paper.

On the other hand, average numbers of tables and figures are steadily growing over the years, with some minor fluctuations post-2022, suggesting these structural trends reflect longer-term disciplinary norms.

## Collaboration and Productivity are Increasing, with Some Help from LLMs

The emergence of LLMs may have reshaped the social dynamics of scientific production, specifically collaboration, team formation, and author productivity. By lowering transaction costs associated with cross-lingual communication, joint manuscript drafting, and editorial revisions, LLMs reduce friction in international co-authorship. This mechanism is particularly salient for researchers in non-English-speaking regions, where language barriers have historically limited participation in scientific discourse.

To test whether the post-2022 period marks a structural inflection in collaborative team formation, we analyzed macro-level trajectories in average co-author team sizes across five geographic regions (North America, Europe, Asia, Other, and Intercontinental collaborations) across all five disciplines. See Supplementary Material for a discussion on how paper origins were identified.

While baseline levels of co-authorship vary substantially by region and discipline, all five disciplines exhibit a well-established, long-term upward trend in average team size prior to 2022 (Figure 6, top row). The post-2022 period shows structural continuity along these pre-existing trajectories. Rather than initiating a new wave of global team expansion, the public availability of LLMs coincided with the continuation of long-term structural shifts in global scientific connectivity. An interesting exception is the post-2022 acceleration of the number of coauthors in Computer Science papers coming from Asia.

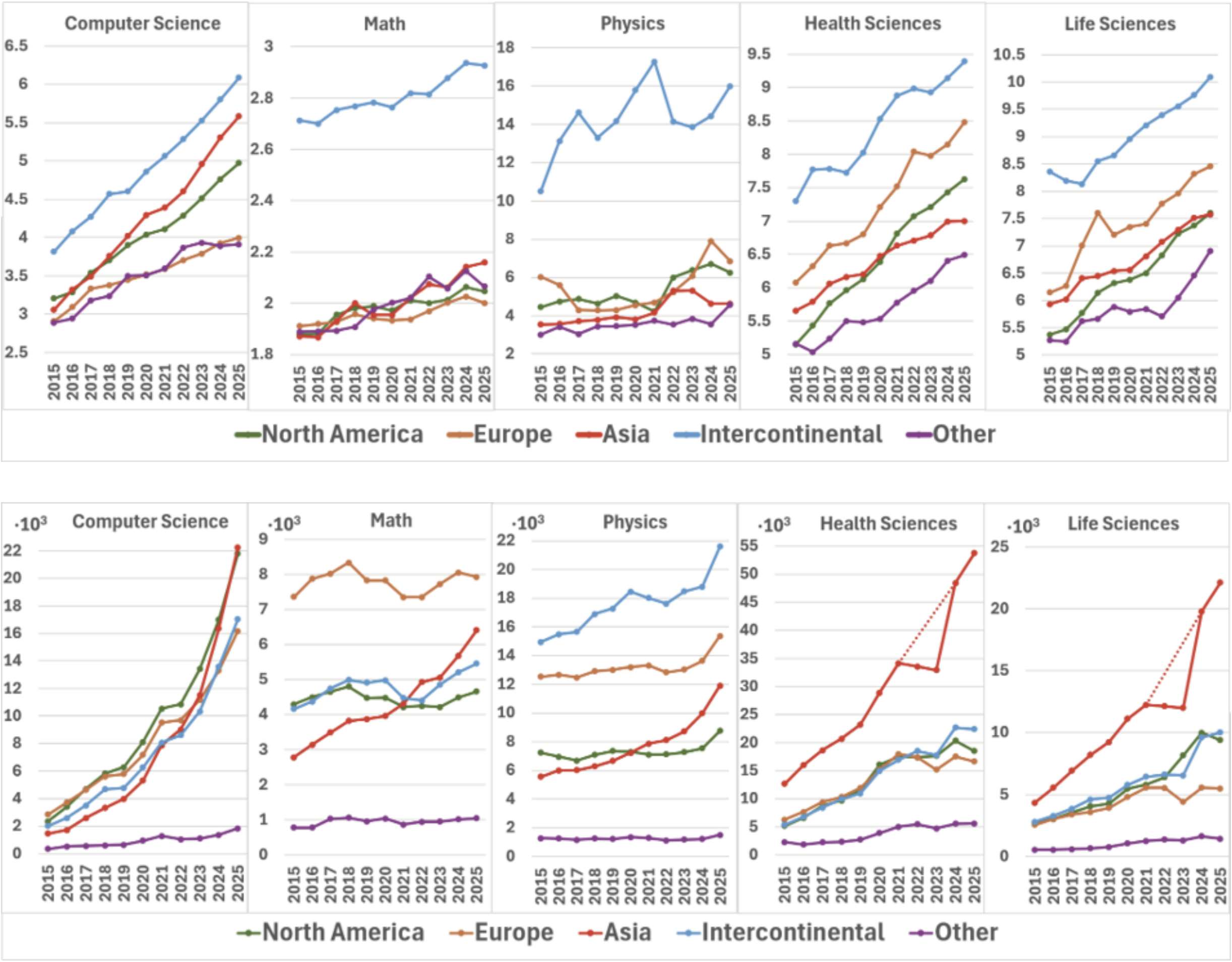

**Figure 6: Social trends in scientific literature.** (Top) average number of paper coauthors; (bottom) total number of papers per their geographical origin. In the graphs of Health and Life Sciences for Asia, we connected the 2021 and 2024 marks with a dotted line, to demonstrate the practically linear trend, disrupted during the COVID pandemic.

We also examined trends in research productivity (in terms of overall number of papers) by region. LLMs have been associated with increased productivity [1]. Figure 6 (bottom row) shows evidence confirming the prior findings. Specifically, the established productivity growth in Computer Science and Physics appears amplified after 2022, which we associate with the emergence of LLMs. In Math, however, the growth is marginal outside Asia (where the growth is very strong since 2015 and even slightly amplified post-2022).

A distinctive growth pattern is detected in Health and Life Sciences. There appears to be a profound inflection point in 2023, especially visible in Asia. However, this dramatic productivity leap is most probably caused by a significant slowdown in 2022, which in turn is likely a result of the COVID-time disruption in biomedical research. Thus, while the inflection point is apparent, its likely cause is not the introduction of LLMs.

## Discussion

Whether the public introduction of Large Language Models in late 2022 marks a structural watershed in the scientific enterprise is a central question for researchers, journal editors, funding agencies, and institutional leaders. Although concerns have been raised about AI tools diminishing the quality of scientific thought, our comprehensive, multi-dimensional analysis reveals that the post-2022 period exhibits structural continuity with previously established evolutionary trajectories in scientific language, epistemology, paper organization, and research collaboration. Across multiple metrics, we found little evidence of the AI paradox at the macro level: despite the productivity benefits realized by individual scholars [2,5], the measured characteristics of scientific writing show little sign of broad post-2022 deterioration.

Our follow-up analyses suggest that the observed trajectory shifts do not provide evidence of LLM-driven epistemic convergence. For instance, the decrease in topic diversity observed in Computer Science long predated 2022 and simply continued along its historical baseline, while marginally slowing down instead of accelerating. Similarly, the apparent 2022 inflection of topic diversity in Health Sciences and Life Sciences largely reflects the standardization of article formats. Filtering boilerplate keyphrases removed most of the decline, and the residual decline likely comes from boilerplate we missed, since some keyphrases used in boilerplate may also be used in research text.

These findings highlight the methodological importance of evaluating the introduction of LLMs through comprehensive frameworks calibrated to pre-existing historical baselines. Isolated signals, such as the surge of specific AI-favored keywords, can easily be misread as signs of systemic transformation. Our full-text analysis reveals a marked contrast: word frequency shifts were not accompanied by comparable breaks in aggregate lexical diversity or the density of core scientific terminology. This continuity suggests that, so far, the emergence of LLMs has not been accompanied by a broad transformation in the observable attributes of scientific writing.

We argue that policy is the right instrument for managing the tension between LLMs and established scientific practices. LLMs are an emerging technology with epistemological implications that are being adopted gradually. Much like science itself, LLM adoption has a trajectory. And much like in science, LLM adoption trajectory can be effectively governed. Presently, policy lags behind both the incentives for researchers to use LLMs [2], and evidence that they are doing so [1,5].

Concerns over AI slop [4], hallucinated references [30,31], questionable ethical practices [32], and reduced innovation [7] may temper the productivity and visibility gains of LLMs, but they have not slowed LLM adoption in scientific research. Attempts to address these problems are already in place as some journals are requiring authors to either affirm that LLMs were not used in manuscript preparation or explain in detail where LLMs were used. Scholarly organizations have also prohibited or greatly restricted the use of LLMs in the peer review process. Nonetheless, LLMs may insidiously weaken guardrails by increasing productivity and flooding publication outlets with paper submissions, and by generating superficial, low-quality research. Our findings suggest that, so far, science has not changed course dramatically. However, the speed with which LLMs are penetrating the research process is concerning. We suggest that future studies use comprehensive frameworks such as ours to better understand LLMs' effects on science and scientists in order to guide effective policy decisions.

# Supplementary Material

## Dataset

We used arXiv's API to download all 2.4 million manuscripts submitted to arXiv between 1991 and 2025. Based on arXiv's paper categorization, we split the dataset to five coarse-grained categories: Computer Science (CS), Math, Physics [33], Other (none of the above), and Interdisciplinary (a combination of the above). A typical manuscript stored in arXiv corresponds in size and content to a journal / conference paper (hence we use the term "paper"), however, some manuscripts are much longer (e.g., a 1185-page book [34]).

In addition, we used a web service API to retrieve 1.7 million full-text papers from PubMed Central (PMC). By querying PMC's Open Access subset, we programmatically downloaded PDF files of papers available under licenses permitting reuse. Since PMC does not natively provide paper categorization, we utilized OpenAlex API called with DOIs of retrieved PMC papers, to obtain their “domain” - which is the highest level category. The two largest domains in our PMC dataset are Health and Life Sciences.

For our analysis, we focus on five disciplines whose scale and longitudinal coverage is sufficient to measure research trajectories reliably (tens of thousands of papers annually in each discipline): Computer Science, Math, and Physics from arXiv as well as Health Sciences and Life Sciences from PMC. We note that arXiv’s category Other is too small while category Interdisciplinary is not distinctive enough. Because CS was too small in the early years of arXiv’s lifecycle, and because those years are less relevant in the context of LLM introduction, we chose to analyze only the past decade, from 2015 to 2025, inclusive. This choice results in a dataset of approximately 2.5 million research papers in their full text form (440k CS, 280k Math, 530k Physics, 890k Health Sciences, and 330k Life Sciences papers).

We used Python’s pymupdf library for extracting text from paper PDFs. We tokenized the texts into words (sequences of alphanumeric characters). For arXiv data, we ended up with 14.8 million unique words, about 10% of which are mathematical notation. We then utilized PaperBasket [26] infrastructure to extract scientific terminology from paper texts. PaperBasket maintains a terminological vocabulary of 20.9 million terms that originate in Wikipedia. Those terms are names of Wikipedia pages and sections within Wikipedia pages, cleaned off various Wikipedia artifacts and augmented with their plural forms. We used Python’s trie data structure to detect those terms in our paper repository. Almost 3.9 million unique terms are detected. Note that the original set of Wikipedia page and section names is comprehensive and not only science-related, but the subset of terms detected in scientific literature is in its vast majority scientific.

## Computation of Topic Diversity

First, let us define topic diversity. We represent each paper as a probability distribution over its terms (i.e. words or phrases, such as scientific terminology). The probability mass of each term is the normalized count of its occurrences in the paper. Both word distributions [35] and terminology distributions [36] capture the paper's topic to a large extent.

**Definition 1:** We define a ***pairwise distance*** between two papers as the Jensen-Shannon (JS) distance between two probability distributions each corresponding to one of the two papers.

JS distance is a square root of JS divergence. JS divergence is widely used in literature for measuring distances between probability distributions in general [37], and language distributions in particular [38]. We choose to use JS distance over JS divergence because the JS distance is a true metric [29]. The range of JS distance values is [0,1].

To compute the JS distance between two papers, we adjust both distributions to have an identical domain which is the union of both papers' term sets. In practice, for a pair of distributions, we pad each distribution with zeros for terms that are absent in its domain but present in the other domain.

**Definition 2:** We define ***topic diversity*** of a paper repository as a mean of pairwise distances of papers in the repository.

Intuitively, a repository of research papers has a low topic diversity whenever pairwise distances between papers are small on average, meaning that papers are topically

“close” to each other and the entire repository is “tight”, or homogeneous. For a typical repository of scientific papers, however, the absolute values of topic diversity are close to 1 because two randomly chosen papers hardly overlap in their words and even less so in their scientific terminology.

Computing topic diversity can come at a high computational cost. For example, for a repository of 100,000 papers, there are about 5 billion unique paper pairs. It is impractical to compute pairwise distances of all paper pairs, and it is not necessary. To establish statistical significance of our results, it is enough to compute the mean of a *sample* of pairwise distances, where pairs of papers in the sample are selected uniformly at random, and the sample size is chosen so that the standard error of the mean is small enough. For word distributions, we used the annual sample of 10,000 pairs which was enough for the standard error of the mean to become very small (below 0.001). For terminology distributions, we used 100,000 pairs, so the standard error of the mean went well below 0.0001. The same sampling framework provides the basis for the non-parametric bootstrap analysis described in the Robustness Analysis below.

## Paper Origin Identification

We are interested in determining where most of the research comes from, and whether there is a qualitative change in research geography over the past decade. Unfortunately, neither arXiv nor PMC record author affiliations or their country of origin in the paper’s metadata. It becomes a challenge to infer the research geography from the texts of papers themselves. Fortunately, for many papers, the country of origin is listed in their headers (which we define as paper portions from the start up to the abstract). Most papers in arXiv (but not in PMC) have their abstracts stored in paper metadata, so there is a distinctive feature defining the header of the paper. In some papers, however, author affiliations appear at the very end of the paper's text, after references. In some other papers, the affiliation information is in a first page's footnote, which may blend together with the text of the introduction. Even when the affiliations are in the paper's header, the country of affiliation is not easy to extract reliably. Examples of confusing cases are Jordan, which can be either a person's name or a Middle Eastern country, and Georgia, which can be either a US State or a European country.

Given this complexity, we chose a cascade approach to extracting paper countries of origin from papers' texts. First, we noticed that email addresses are not difficult to detect - and they virtually always correspond to paper authors, while being easily traceable to the author's country of affiliation. We extracted email addresses and traced them to countries. If this step failed (some papers don't list email addresses, or some email domains are not country-specific), we extracted countries from paper headers. In a rare case of the previous stages' failure, we extracted country names from everywhere else in the paper. We integrated the results and fixed some inconsistencies. At the end, 92% of the papers were associated with their countries of origin.

We traced paper countries of origin to their continents, and noticed that the majority of research comes from North America, Europe, and Asia, while the rest of the world is heavily underrepresented. We united the three remaining continents into the category of Other. Also, we identified a very common case when the authors of a paper are based in multiple continents. For those papers, we created a special category "Intercontinental".

## AI-favored word usage in abstracts and paper bodies

While our paper analyzes full texts of research papers, many related studies were conducted on paper abstracts [2,3,8-10,12,15,17,21]. A paper abstract is the part of a paper most likely drafted or polished with assistance, so it is reasonable to assume that AI-favored words would be used in abstracts more commonly than in paper bodies. Over the arXiv subset of our dataset, we measured frequencies of surge words in abstracts and paper bodies. Both rose after 2022 with about the same rate. In abstracts, the rate went from 0.240% in 2015 to 0.548% in 2025, a rise of 128%. In paper bodies, it went from 0.110% to 0.254%, a rise of 131% (Figure 7, top). Surge words are about twice as frequently used in abstracts, and that gap is significant in every year (paired $t$-test on per-article ratios: $t$ = 65 to 161, all $p < 10^{-300}$). But the gap does not grow over time. It averages 2.17 and stays between 2.11 and 2.21 across the eleven years (Figure 7, bottom). The abstract/body usage ratio of decline words stays near the same too, 0.84 to 0.95. In summary, the increase in AI-favored vocabulary is not confined to paper abstracts.

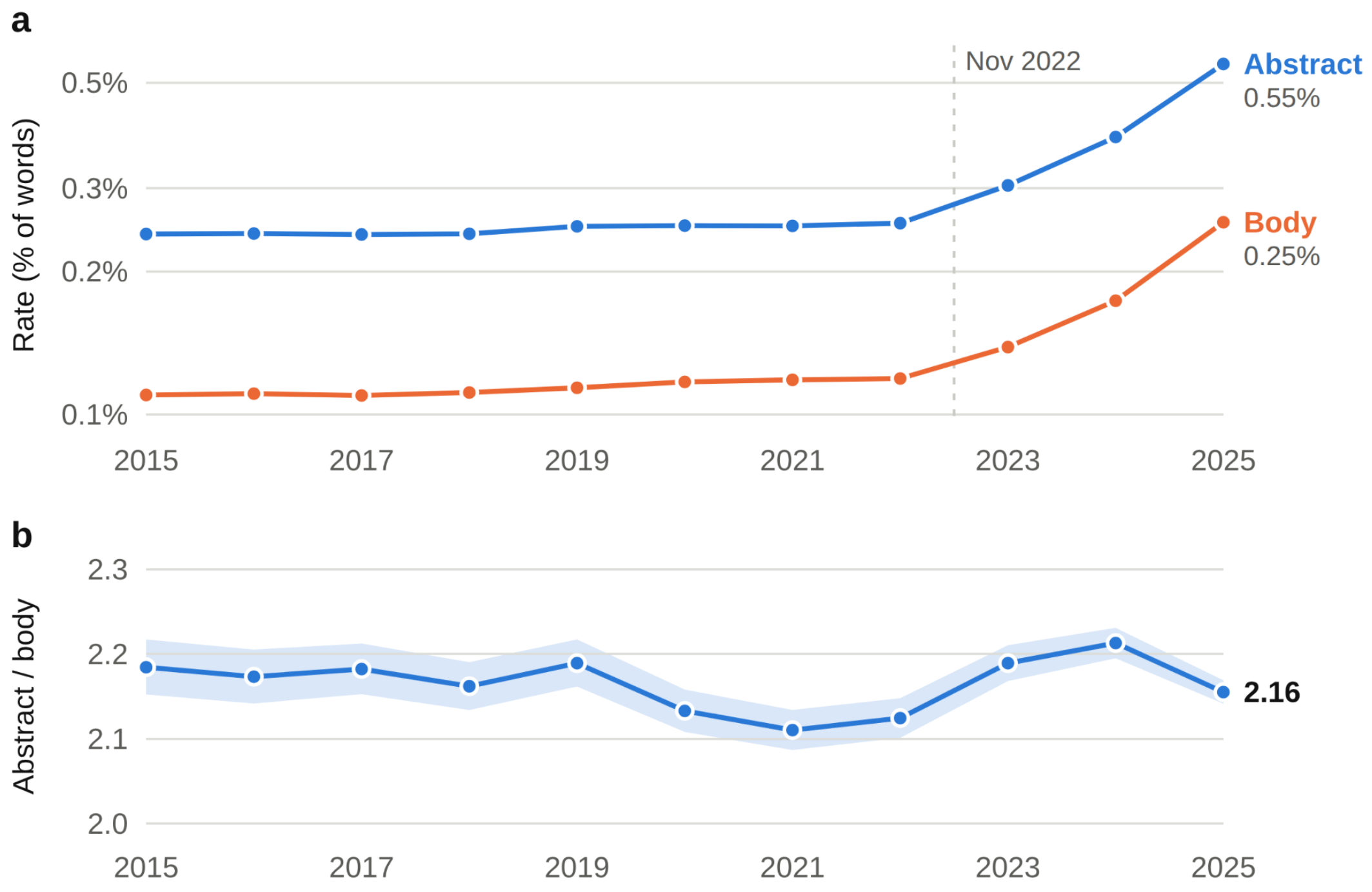

**Figure 7: The post-2022 increase is distributed through the article, not confined to abstracts**. (top) Rate of the 20 surge words per year in abstracts (blue) and article bodies (orange), plotted on a logarithmic axis so that equal proportional change appears as equal vertical displacement. (bottom) The abstract/body ratio of AI-favored word usage.

## Statistical Analysis

To evaluate potential structural inflections in scientific publishing trends following the emergence of LLMs at the end of 2022, we used Ordinary Least Squares (OLS) to fit a segmented linear regression model with a continuous knot fixed at 2022 [39,40]. We evaluated the annual evolution of metrics proposed in this paper across Computer Science, Mathematics, Physics, Health Sciences, and Life Sciences. For each discipline, time was decomposed into a baseline trend $t_{baseline} = Year - 2015$ and a post-knot spline variable $t_{knot} = \max(0, Year - 2022)$. Under this parameterization, $\beta_1$ estimates the baseline annual rate of change during the pre-break period (2015–2022), $\beta_2$ isolates the post-2022 slope differential, and the composite term $\beta_1 + \beta_2$ measures the net trajectory following the 2022 inflection point.

Statistical inference was conducted to evaluate the null hypothesis ($H_0$: $\beta_2 = 0$), which states that the post-2022 slope shift does not significantly differ from zero. Failure to reject this hypothesis indicates that the data do not provide sufficient evidence of a change in the trajectory of a given metric. To guard against Type I error inflation resulting from simultaneous hypothesis testing across $K = 5$ disciplinary models, we applied a Bonferroni adjustment [41], establishing a family-wise significance threshold of $\alpha = 0.05/5 = 0.01$. Two-tailed $p$-values were derived alongside 99% Bonferroni-adjusted confidence intervals using Student's $t$-distribution with 8 residual degrees of freedom (11 annual observations minus 3 estimated parameters). Null hypotheses yielding $p < 0.01$ were rejected in favor of a statistically significant post-2022 structural break. Finally, to differentiate quantitative acceleration or deceleration from qualitative trajectory reversals, we assessed directional retention by comparing the sign of the baseline slope $sign(\beta_1)$ with the net post-2022 slope $sign(\beta_1 + \beta_2)$, determining whether a metric's direction remained consistent or flipped.

Residuals ($e_t = y_t - \hat{y}_t$) from each segmented annual regression (2015–2025) were evaluated for temporal dependence using the Durbin–Watson (DW) statistic, lag-1 and lag-2 residual autocorrelations (ACF1 and ACF2), and the Ljung–Box Q-test at lag 1. OLS standard errors were used as the primary basis for inference, with Newey–West heteroskedasticity- and autocorrelation-consistent (HAC) standard errors using a lag of 1 year applied as a sensitivity analysis, when the diagnostics indicated residual autocorrelation. The results below report OLS estimates and inference unless otherwise specified.

## Lexical Diversity (Figure 3, first panel)

| Discipline | Pre-2023 Slope ($\beta_1$) | Post-2022 Slope ($\beta_1$+$\beta_2$) | Slope Change ($\beta_2$) | Two-Tailed p-Value | Bonferroni 99% CI | Significant at α = 0.01? | Same Direction of the Trend? |
|---|---|---|---|---|---|---|---|
| Computer Science | 0.00386 | 0.00633 | 0.00247 | 0.24774 | [-0.00418, +0.00912] | No | Yes (Upward) |
| Math | -0.00057 | -0.00009 | 0.00048 | 0.01201 | [-0.00002, +0.00097] | No | Yes (Downward) |
| Physics | 0.00067 | 0.00302 | 0.00236 | 0.0133 | [-0.00014, +0.00485] | No | Yes (Upward) |
| Health Sciences | -0.00003 | 0.00026 | 0.00029 | 0.66559 | [-0.00188, +0.00246] | No | No (Downward → Upward) |
| Life Sciences | -0.00045 | 0.00149 | 0.00195 | 0.05572 | [-0.00097, +0.00487] | No | No (Downward → Upward) |

There is no statistically significant slope change in any of the disciplines. No autocorrelation was detected.

## Average Vocabulary Size (Figure 3, second panel)

| Discipline | Pre-2023 Slope ($\beta_1$) | Post-2022 Slope ($\beta_1$+$\beta_2$) | Slope Change ($\beta_2$) | Two-Tailed p-Value | Bonferroni 99% CI | Significant at α = 0.01? | Same Direction of the Trend? |
|---|---|---|---|---|---|---|---|
| Computer Science | 52.30175 | 96.1015 | 43.79975 | 0.00186 | [+11.53413, +76.06537] | Yes | Yes (Upward) |
| Math | 10.63514 | 12.22135 | 1.58622 | 0.71761 | [-12.61752, +15.78996] | No | Yes (Upward) |
| Physics | 40.4879 | 27.64617 | -12.8417 | 0.10562 | [-36.46667, +10.78320] | No | Yes (Upward) |
| Health Sciences | 24.38514 | 55.90405 | 31.5189 | 0.03654 | [-10.66672, +73.70453] | No | Yes (Upward) |
| Life Sciences | 34.27124 | 87.08679 | 52.81554 | 0.022 | [-9.70946, +115.34055] | No | Yes (Upward) |

All trends are upward. There are no inflection points in 2022, except for Computer Science that showed a dramatic growth pre-2023 which appears accelerating even further post-2022. While Computer Science exhibits borderline positive residual autocorrelation ($DW = 0.847, ACF_1 =+ 0.508, p = 0.0548$), adjusting for temporal persistence via Newey-West HAC standard errors leaves the post-2022 trajectory inflection highly significant ($p_{HAC} < 0.0001$), matching the baseline OLS conclusion.

## Average Paper Length (Figure 3, third panel)

| Discipline | Pre-2023 Slope ($\beta_1$) | Post-2022 Slope ($\beta_1$+$\beta_2$) | Slope Change ($\beta_2$) | Two-Tailed p-Value | Bonferroni 99% CI | Significant at α = 0.01? | Same Direction of the Trend? |
|---|---|---|---|---|---|---|---|
| Computer Science | 162.7903 | 231.256 | 68.46573 | 0.64569 | [-412.51044, +549.44191] | No | Yes (Upward) |
| Math | 265.4455 | 63.97985 | -201.466 | 0.07655 | [-534.04955, +131.11820] | No | Yes (Upward) |
| Physics | 305.8369 | -23.9908 | -329.828 | 0.00472 | [-615.54190, -44.11354] | Yes | No (Upward -> Downward) |
| Health Sciences | 137.2639 | 330.6638 | 193.3999 | 0.00768 | [+9.82609, +376.97374] | Yes | Yes (Upward) |
| Life Sciences | 229.8258 | 538.8535 | 309.0277 | 0.02136 | [-54.32579, +672.38112] | No | Yes (Upward) |

All trends besides Physics are upward. In Physics, the upward trend halts in 2025, causing a statistically significant decrease. We associate this decrease with a correction that Physics experienced after a dramatic COVID-period growth. In Health Sciences, the upward trend appears accelerating post-2022. Computer Science displays statistically significant residual autocorrelation ($DW = 0.688, ACF_1 =+ 0.547, p_{Ljung-Box} = 0.0386$), however, adjusting for this temporal dependence via Newey-West HAC standard errors confirms that the post-2022 slope change remains non-significant ($p_{HAC} = 0.52680$).

## Scientific Terminology (Figure 3, fourth panel)

| Discipline | Pre-2023 Slope ($\beta_1$) | Post-2022 Slope ($\beta_1+\beta_2$) | Slope Change ($\beta_2$) | Two-Tailed p-Value | Bonferroni 99% CI | Significant at $\alpha = 0.01$? | Same Direction of the Trend? |
|---|---|---|---|---|---|---|---|
| Computer Science | 0.00344 | 0.00351 | 0.00007 | 0.86183 | [-0.00117, +0.00130] | No | Yes (Upward) |
| Math | 0.00057 | 0.00095 | 0.00038 | 0.30354 | [-0.00078, +0.00154] | No | Yes (Upward) |
| Physics | 0.00267 | 0.00159 | -0.00109 | 0.00966 | [-0.00217, -0.00001] | Yes | Yes (Upward) |
| Health Sciences | 0.00147 | 0.00423 | 0.00276 | 0.01549 | [-0.00026, +0.00578] | No | Yes (Upward) |
| Life Sciences | 0.00178 | 0.00319 | 0.00141 | 0.30921 | [-0.00294, +0.00576] | No | Yes (Upward) |

Only the Physics trend slowed down marginally (low $\beta_2$), all the rest stayed steadily upward. No autocorrelation was detected.

## Topic Diversity over Word Distributions (Figure 4, left panel)

| Discipline | Pre-2023 Slope ($\beta_1$) | Post-2022 Slope ($\beta_1+\beta_2$) | Slope Change ($\beta_2$) | Two-Tailed p-Value | Bonferroni 99% CI | Significant at $\alpha = 0.01$? | Same Direction of the Trend? |
|---|---|---|---|---|---|---|---|
| Computer Science | -0.00329 | -0.00247 | 0.00082 | 0.18045 | [-0.00106, +0.00270] | No | Yes (Downward) |
| Math | -0.00064 | 0.00038 | 0.00103 | 0.14804 | [-0.00113, +0.00318] | No | No (Downward -> Upward) |
| Physics | -0.00105 | 0.00032 | 0.00137 | 0.02928 | [-0.00036, +0.00310] | No | No (Downward -> Upward) |
| Health Sciences | -0.00118 | -0.00496 | -0.00378 | 0.00799 | [-0.00740, -0.00016] | Yes | Yes (Downward) |
| Life Sciences | -0.00125 | -0.00566 | -0.00441 | 0.00045 | [-0.00700, -0.00181] | Yes | Yes (Downward) |

In Computer Science, Math, and Physics, the trend stays the same (for CS, it is downward, while for Math and Physics it is practically flat). In Health and Life Sciences the downward trend significantly accelerates, though the absolute values of the slope change ($\beta_2$) are small. As we discussed in the paper, the downward trend is primarily caused by intensifying requirements of paper format unification, rather than by topic homogenization. No autocorrelation was detected.

## Topic Diversity over Scientific Terminology Distributions (Figure 4, right panel, solid lines)

| Discipline | Pre-2023 Slope ($\beta_1$) | Post-2022 Slope ($\beta_1+\beta_2$) | Slope Change ($\beta_2$) | Two-Tailed p-Value | Bonferroni 99% CI | Significant at $\alpha = 0.01$? | Same Direction of the Trend? |
|---|---|---|---|---|---|---|---|
| Computer Science | -0.00186 | -0.00113 | 0.00074 | 0.00885 | [+0.00002, +0.00146] | Yes | Yes (Downward) |
| Math | -0.00016 | 0.00025 | 0.00041 | 0.01274 | [-0.00002, +0.00085] | No | No (Downward -> Upward) |
| Physics | -0.00027 | -0.00012 | 0.00015 | 0.24172 | [-0.00025, +0.00054] | No | Yes (Downward) |
| Health Sciences | -0.00018 | -0.00066 | -0.00047 | 0.03077 | [−0.00107, +0.00013] | No | Yes (Downward) |
| Life Sciences | -0.00015 | -0.00052 | -0.00037 | 0.35756 | [−0.00164, +0.00090] | No | Yes (Downward) |

After we filtered out formatting artifacts of Health and Life Sciences papers, the five disciplines showed similar results: the slope change for all disciplines is negligibly small (although for CS it ends up being statistically significant, with the lower bound of the 99% confidence interval hitting 0.00002). See robustness tests below. Residual diagnostics indicated evidence of lag-1 autocorrelation in the Life Sciences model, however, inference was unchanged when using HAC standard errors, with the estimated post-2022 slope change remaining non-significant and unchanged in its direction.

## Average References per Paper (Figure 5, panel e)

| Discipline | Pre-2023 Slope ($\beta_1$) | Post-2022 Slope ($\beta_1$+$\beta_2$) | Slope Change ($\beta_2$) | Two-Tailed p-Value | Bonferroni 99% CI | Significant at α = 0.01? | Same Direction of the Trend? |
|---|---|---|---|---|---|---|---|
| Computer Science | 2.05646 | 0.40606 | -1.6504 | 0.00469 | [-3.07823, -0.22257] | Yes | Yes (Upward) |
| Math | 0.52154 | 0.07273 | -0.44881 | 0.01016 | [-0.89902, +0.00140] | No | Yes (Upward) |
| Physics | 1.54304 | 0.23788 | -1.30516 | 0.02374 | [-2.87754, +0.26723] | No | Yes (Upward) |
| Health Sciences | 0.86043 | 2.58939 | 1.72897 | 0.00056 | [+0.67686, +2.78107] | Yes | Yes (Upward) |
| Life Sciences | 1.17316 | 3.06364 | 1.89048 | 0.00935 | [+0.02527, +3.75568] | Yes | Yes (Upward) |

For Computer Science, Math, and Physics, the number of references is plateauing post-2022, following the growth over the COVID period. Only for CS the slowdown is significant, showing signs of standardization of CS paper structure. Health and Life Sciences, to the contrary, experience a significant acceleration of the upward trend, which was set well before 2022. No autocorrelation was detected.

## Average Tables per Paper (Figure 5, panel f)

| Discipline | Pre-2023 Slope ($\beta_1$) | Post-2022 Slope ($\beta_1$+$\beta_2$) | Slope Change ($\beta_2$) | Two-Tailed p-Value | Bonferroni 99% CI | Significant at α = 0.01? | Same Direction of the Trend? |
|---|---|---|---|---|---|---|---|
| Computer Science | 0.27571 | 0.18448 | -0.09123 | 0.09504 | [-0.25297, +0.07051] | No | Yes (Upward) |
| Math | 0.00206 | 0.02154 | 0.01948 | 0.09154 | [-0.01461, +0.05356] | No | Yes (Upward) |
| Physics | 0.01471 | -0.01975 | -0.03446 | 0.01745 | [-0.07319, +0.00427] | No | No (Upward -> Downward) |
| Health Sciences | 0.04892 | 0.16661 | 0.1177 | 0.00468 | [+0.01588, +0.21951] | Yes | Yes (Upward) |
| Life Sciences | 0.07737 | 0.21367 | 0.1363 | 0.00042 | [+0.05714, +0.21546] | Yes | Yes (Upward) |

For all disciplines besides Physics the trend is upward, with some slight (low $\beta_2$) but significant acceleration in Health and Life Sciences. The Physics trend is practically flat. No autocorrelation was detected.

## Average Figures per Paper (Figure 5, panel g)

| Discipline | Pre-2023 Slope ($\beta_1$) | Post-2022 Slope ($\beta_1$+$\beta_2$) | Slope Change ($\beta_2$) | Two-Tailed p-Value | Bonferroni 99% CI | Significant at $\alpha$ = 0.01? | Same Direction of the Trend? |
|---|---|---|---|---|---|---|---|
| Computer Science | 0.1363 | 0.14928 | 0.01298 | 0.83833 | [-0.19373, +0.21970] | No | Yes (Upward) |
| Math | 0.01624 | 0.06554 | 0.0493 | 0.09532 | [-0.03819, +0.13679] | No | Yes (Upward) |
| Physics | 0.12719 | -0.03081 | -0.158 | 0.00252 | [-0.28051, -0.03548] | Yes | No (Upward -> Downward) |
| Health Sciences | 0.07787 | 0.26821 | 0.19034 | 0.00012 | [+0.09866, +0.28202] | Yes | Yes (Upward) |
| Life Sciences | 0.20801 | 0.50564 | 0.29763 | 0.00466 | [+0.04039, +0.55486] | Yes | Yes (Upward) |

All trends are upward, with Computer Science and Math growing slowly and steadily (no inflection point post-2022). For Physics, the upward trend appears to have reached a plateau, and even gets slightly reversed. For Health and Life Sciences, slight post-2022 trend acceleration is visible and statistically significant. No autocorrelation was detected.

## Average Number of Paper Coauthors (Figure 6, top panel)

| Discipline | Paper Origin | Pre-2023 Slope ($\beta_1$) | Post-2022 Slope ($\beta_1$+$\beta_2$) | Slope Change ($\beta_2$) | Two-Tailed p-Value | Bonferroni 99% CI | Significant at $\alpha$ = 0.01? | Same Direction of the Trend? |
|---|---|---|---|---|---|---|---|---|
| Computer Science | North America | 0.15871 | 0.21938 | 0.06067 | 0.019 | [-0.00881, +0.13016] | No | Yes (Upward) |
| | Europe | 0.10224 | 0.0904 | -0.01184 | 0.70304 | [-0.11233, +0.08865] | No | Yes (Upward) |
| | Asia | 0.22433 | 0.31396 | 0.08962 | 0.00314 | [+0.01742, +0.16182] | Yes | Yes (Upward) |
| | Intercontinental | 0.20148 | 0.26813 | 0.06665 | 0.01662 | [-0.00746, +0.14077] | No | Yes (Upward) |
| | Other | 0.12609 | 0.03549 | -0.0906 | 0.098 | [-0.25293, +0.07173] | No | Yes (Upward) |
| Math | North America | 0.01817 | 0.01044 | -0.00774 | 0.52294 | [-0.04660, +0.03113] | No | Yes (Upward) |
| | Europe | 0.00892 | 0.01577 | 0.00685 | 0.46461 | [-0.02307, +0.03676] | No | Yes (Upward) |
| | Asia | 0.02429 | 0.0392 | 0.01491 | 0.3039 | [-0.03062, +0.06044] | No | Yes (Upward) |
| | Intercontinental | 0.01805 | 0.03916 | 0.02111 | 0.03259 | [-0.00634, +0.04855] | No | Yes (Upward) |
| | Other | 0.03075 | 0.01098 | -0.01978 | 0.30233 | [-0.07997, +0.04042] | No | Yes (Upward) |
| Physics | North America | 0.11167 | 0.51172 | 0.40005 | 0.12528 | [-0.38408, +1.18417] | No | Yes (Upward) |
| | Europe | 0.0603 | 0.8938 | 0.83351 | 0.02069 | [-0.13951, +1.80652] | No | Yes (Upward) |
| | Asia | 0.21379 | -0.00629 | -0.22007 | 0.25922 | [-0.82813, +0.38798] | No | No (Upward -> Downward) |
| | Intercontinental | 0.53352 | -0.46498 | -0.99849 | 0.14254 | [-3.05859, +1.06160] | No | No (Upward -> Downward) |
| | Other | 0.08084 | 0.08936 | 0.00852 | 0.91785 | [-0.26002, +0.27705] | No | Yes (Upward) |
| Health Sciences | North America | 0.26612 | 0.2022 | -0.06392 | 0.06038 | [-0.16208, +0.03423] | No | Yes (Upward) |
| | Europe | 0.25357 | 0.21221 | -0.04136 | 0.54903 | [-0.26323, +0.18050] | No | Yes (Upward) |
| | Asia | 0.1518 | 0.09586 | -0.05594 | 0.07468 | [-0.14757, +0.03569] | No | Yes (Upward) |
| | Intercontinental | 0.22844 | 0.13602 | -0.09242 | 0.28488 | [-0.36301, +0.17816] | No | Yes (Upward) |
| | Other | 0.12272 | 0.21663 | 0.09391 | 0.0633 | [-0.05233, +0.24014] | No | Yes (Upward) |
| Life Sciences | North America | 0.2125 | 0.2615 | 0.049 | 0.31566 | [-0.10461, +0.20261] | No | Yes (Upward) |
| | Europe | 0.20767 | 0.21846 | 0.01079 | 0.93643 | [-0.42896, +0.45053] | No | Yes (Upward) |
| | Asia | 0.15622 | 0.20653 | 0.05031 | 0.30251 | [-0.10287, +0.20348] | No | Yes (Upward) |
| | Intercontinental | 0.17742 | 0.25017 | 0.07274 | 0.35142 | [-0.17394, +0.31943] | No | Yes (Upward) |
| | Other | 0.07193 | 0.31903 | 0.2471 | 0.00758 | [+0.01311, +0.48109] | Yes | Yes (Upward) |

Practically all the trends are upward and the change of slope post-2022 is insignificant. Out of two exceptions, one is remarkable: growth acceleration in CS paper coauthorship teams in Asia. The category “Other” was found autocorrelating for CS and Physics, and while for Physics the HAC p-value remained above 0.01, for CS it went under 0.01.

## Paper Volumes per Geographical Origin (Figure 6, bottom panel)

| Discipline | Paper Origin | Pre-2023 Slope ($\beta_1$) | Post-2022 Slope ($\beta_1$ + $\beta_2$) | Slope Change ($\beta_2$) | Two-Tailed p-Value | Bonferroni 99% CI | Significant at $\alpha$ = 0.01? | Same Direction of the Trend? |
|---|---|---|---|---|---|---|---|---|
| Computer Science | North America | 1202.31 | 3501 | 2298.69 | 0.00005 | [+1304.87109, +3292.50986] | Yes | Yes (Upward) |
| | Europe | 974.7136 | 2098.848 | 1124.135 | 0.00208 | [+280.84954, +1967.42030] | Yes | Yes (Upward) |
| | Asia | 1030.469 | 4536.061 | 3505.591 | 0.00003 | [+2135.86377, +4875.31877] | Yes | Yes (Upward) |
| | Intercontinental | 945.316 | 2760.364 | 1815.048 | 0.00008 | [+992.71812, +2637.37712] | Yes | Yes (Upward) |
| | Other | 104.3745 | 202.1364 | 97.7619 | 0.20307 | [-138.86359, +334.38740] | No | Yes (Upward) |
| Math | North America | -51.5065 | 147.1364 | 198.6429 | 0.10421 | [-165.05308, +562.33879] | No | No (Downward -> Upward) |
| | Europe | -63.7406 | 215.9697 | 279.7103 | 0.15746 | [-322.04119, +881.46182] | No | No (Downward -> Upward) |
| | Asia | 252.0491 | 554.803 | 302.754 | 0.00666 | [+23.12384, +582.38410] | Yes | Yes (Upward) |
| | Intercontinental | 11.47799 | 250.3788 | 238.9008 | 0.08829 | [-174.21235, +652.01393] | No | Yes (Upward) |
| | Other | 17.77273 | 12.77273 | -5 | 0.91569 | [-158.55668, +148.55668] | No | Yes (Upward) |
| Physics | North America | -1.03932 | 445.2424 | 446.2818 | 0.0086 | [+13.15258, +879.41091] | Yes | No (Downward -> Upward) |
| | Europe | 34.80483 | 623.5152 | 588.7103 | 0.01556 | [-56.65191, +1234.07255] | No | Yes (Upward) |
| | Asia | 359.4794 | 1182.182 | 822.7024 | 0.00074 | [+300.14381, +1345.26095] | Yes | Yes (Upward) |
| | Interconti nental | 441.3121 | 789.5303 | 348.2183 | 0.3236 | [-762.58772, +1459.02423] | No | Yes (Upward) |
| | Other | -6.97114 | 66.56061 | 73.53175 | 0.12884 | [-72.17003, +219.23353] | No | No (Downward -> Upward) |
| Health Sciences | North America | 1937.205 | 262.5303 | -1674.67 | 0.01325 | [-3448.83682, +99.48762] | No | Yes (Upward) |
| | Europe | 1669.681 | -436.045 | -2105.73 | 0.00436 | [-3903.85003, -307.60235] | Yes | No (Upward -> Downward) |
| | Asia | 2857.76 | 6365.045 | 3507.286 | 0.03068 | [-985.50965, +8000.08107] | No | Yes (Upward) |
| | Interconti nental | 1883.408 | 1598.182 | -285.226 | 0.59598 | [-2018.78958, +1448.33720] | No | Yes (Upward) |
| | Other | 496.952 | 220.4242 | -276.528 | 0.31173 | [-1135.86065, +582.80509] | No | Yes (Upward) |
| Life Sciences | North America | 598.618 | 1133.273 | 534.6548 | 0.0878 | [-388.17262, +1457.48214] | No | Yes (Upward) |
| | Europe | 412.4499 | -26.0303 | -438.48 | 0.07295 | [-1151.43181, +274.47149] | No | No (Upward -> Downward) |
| | Asia | 1049.47 | 3181.788 | 2132.317 | 0.01094 | [-38.88740, +4303.52232] | No | Yes (Upward) |
| | Interconti nental | 531.9365 | 1179.667 | 647.7302 | 0.03124 | [-185.71037, +1481.17069] | No | Yes (Upward) |
| | Other | 124.9928 | 74.98485 | -50.0079 | 0.40987 | [-242.97266, +142.95678] | No | Yes (Upward) |

The majority of trends are upward. Most Computer Science trends significantly accelerate post-2022. Paper submissions originating in Asia accelerate for Math and Physics, as well as for CS. In addition, the pre-2023 flat trend in Physics paper volumes coming from North America switched to a significant growth, while Health and Life Sciences paper volume growth coming from Europe appears to halt. The category

“Intercontinental” was found borderline autocorrelating for Math, however its HAC p-value stayed above 0.01.

## Sensitivity and Robustness Analyses

We performed sensitivity and robustness analyses on the topic diversity results when papers are represented as terminology distributions (Figure 4, right panel, solid lines). The steady decrease in Computer Science topic diversity while all the other four disciplines maintain their topic diversity practically unchanged is our key result, which is why we chose to examine it in greater detail.

Our primary analysis described above employed segmented linear regression to evaluate whether the temporal trajectory changed after a prespecified breakpoint in 2022. This modeling framework estimates the pre-break trend ($\beta_1$), the post-break slope change ($\beta_2$), and the net post-break trajectory ($\beta_1+\beta_2$), serving as a standard approach for evaluating changes in longitudinal trajectories [22, 42].

### Bayesian Evidence and Non-Inferiority Testing

To test whether topic diversity declined faster after 2022 than its earlier trajectory would predict, we fitted each discipline’s annual mean pairwise Jensen–Shannon (JS) distance series (Figure 4, right panel, solid lines) with a linear trend and a change in slope after 2022 [22, 42]. The slope change, $\beta_2$, is the difference between the pre-2023 and post-2022 slopes. The implied 2025 departure of the fitted trajectory from the extrapolated pre-2023 trend equals $3\beta_2$ when time is measured in years from 2015.

Because a non-significant slope change does not establish that the change is negligible [43], we used a one-sided non-inferiority test [44]: for a margin ε, we asked whether the 2025 departure exceeds −ε, using the lower one-sided 99% confidence bound (familywise one-sided $\alpha = 0.05$, Bonferroni-adjusted to 0.01 per discipline). We report the smallest margin each discipline passes. Inference used ordinary least-squares (OLS) standard errors with 8 degrees of freedom; for Life Sciences, where residual autocorrelation was detected, we added a HAC correction (Newey–West, lag 1) as a sensitivity check.

We also computed a Bayes factor ($BF_{01}$) [45] comparing no slope change ($H_0$: $\beta_2 = 0$) with an accelerated decline ($H_1$: $\beta_2 < 0$). It was computed from the estimated $\beta_2$ and its standard error, using a t likelihood with the model's residual degrees of freedom and a half-normal prior on the magnitude of the 2025 departure ($3\beta_2$) with scale 0.001 JS. Because $H_1$ includes only accelerating declines, a positive slope change favors $H_0$ relative to this directional alternative, although it does not establish that the slope change is zero. Values above 3 favor no acceleration; values below 1/3 favor accelerated decline. We examined sensitivity to the margin and prior scale, and repeated the analysis without 2025 to assess its influence.

All five series are approximately linear over the study period (residual SD 0.0003–0.0008 JS from straight-line fits, largest in Life Sciences) - for context, Computer Science declined by 0.016 JS over the decade. The estimated 2025 departure was positive in Computer Science, Math, and Physics, indicating a slower decline, though for Physics it was not distinguishable from zero. Computer Science and Math passed non-inferiority even at a zero margin. Their $BF_{01}$ values were 3.4 and 5.2, respectively, favoring no acceleration over accelerated decline. Physics passed non-inferiority only at margins of 0.00058 JS or larger; its $BF_{01}$ was 5.3. Health Sciences showed an estimated acceleration (departure = −0.00141 JS; $BF_{01}$ = 0.14, about 7:1 in favor of accelerated decline). For Life Sciences the evidence was inconclusive: the data could not rule out an accelerated decline of up to 0.00441 JS. The table reports OLS-based results; for Life Sciences, HAC standard errors gave a slightly smaller minimum margin (0.0038 JS) and did not change the conclusion.

| **Discipline** | **2025 departure** | **Lower bound (one-sided 99%)** | **Margin** | **$BF_{01}$** | **Margin, 2025 omitted** | **$BF_{01}$, 2025 omitted** |
|---|---|---|---|---|---|---|
| Computer Science | +0.00221 | +0.00035 | 0 | 3.4 | 0.00077 | 2.4 |
| Math | +0.00124 | +0.00012 | 0 | 5.2 | 0.00056 | 3.5 |
| Physics | +0.00045 | −0.00058 | 0.00058 | 5.3 | 0.00144 | 1.4 |
| Health Sciences | −0.00141 | −0.00298 | 0.00298 | 0.14 | 0.00167 | 1.7 |
| Life Sciences | −0.00111 | −0.00441 | 0.00441 | 0.72 | 0.00452 | 1.3 |

The conclusions for Computer Science and Math held across all margins examined (0.0002–0.005 JS). Their $BF_{01}$ values exceeded 3 at prior scales of 0.001 JS or larger for Computer Science and 0.0006 JS or larger for Math. At the largest scale (0.005 JS) they reached 14 and 24, respectively, but fell to 1.9 and 2.5 at the smallest scale (0.0004 JS), so this evidence is moderate and prior-dependent. Physics passed only at margins of about 0.0006 JS or larger, and its slope change reversed sign when 2025 was omitted. The Health Sciences result also depended on 2025: without it, the evidence for accelerated decline disappeared ($BF_{01}$ = 1.7), so it should not be read as a sustained post-2022 change. Omitting 2025 leaves two post-break observations, so this is a diagnostic, not an equally informative analysis. More generally, with only three post-2022 observations, these analyses bound a faster decline rather than rule it out.

## Non-Parametric Block Bootstrap

To complement models fitted on annual aggregated means, we conducted a year-stratified non-parametric block bootstrap (with 10,000 iterations) directly on the unaggregated data (see Computation of Topic Diversity), comprising 5.5 million pairwise JS distance observations across 5 disciplines and 11 years. In each bootstrap iteration, 100,000 pairwise JS distances were randomly resampled with replacement from the empirical pool of each discipline-year. Synthetic annual means were calculated from these resampled draws to form an 11-year time series, over which the segmented regression model was re-estimated. Empirical 99% confidence intervals were derived from the 0.5th and 99.5th percentiles of the empirical distribution.

| **Discipline** | **Pre-2023 Slope ($\beta_1$)** | **Post-2022 Slope ($\beta_1$+$\beta_2$)** | **Slope Change ($\beta_2$)** | **Bootstrap p-Value** | **Bonferroni 99% Empirical CI** | **Significant at α = 0.01?** | **Same Direction of the Trend?** |
|---|---|---|---|---|---|---|---|
| Computer Science | -0.00185 | -0.00114 | 0.00071 | < 0.0002 | [+0.00065, +0.00077] | Yes | Yes (Downward) |
| Math | -0.00013 | 0.00016 | 0.00029 | < 0.0002 | [+0.00023, +0.00035] | Yes | No (Downward → Upward) |
| Physics | -0.00028 | -0.00014 | 0.00014 | < 0.0002 | [+0.00009, +0.00018] | Yes | Yes (Downward) |
| Health Sciences | -0.00018 | -0.00069 | -0.00052 | < 0.0002 | [-0.00056, -0.00047] | Yes | Yes (Downward) |
| Life Sciences | -0.00014 | -0.00055 | -0.00041 | < 0.0002 | [-0.00046, -0.00036] | Yes | Yes (Downward) |

The parameter point estimates obtained from the unaggregated bootstrap align closely with those derived from the 11 annual averages, with post-2022 slope changes ($\beta_2$) becoming even smaller in absolute magnitude and empirical confidence intervals becoming tighter. All slope changes yield $p < 0.0002$ simply because the massive sample size ($N \approx 5.5 million$) compresses the standard errors of the mean to $\sim 10^{-5}$. Placebo tests placing arbitrary knots at pre-LLM years (e.g., 2019) similarly produced $p < 0.0002$ across all disciplines, confirming that this statistical significance reflects extreme sample size power rather than a unique structural inflection in 2022. Across all disciplines, the estimated annual slope shift remains substantively imperceptible (a 0.06% shift on a baseline JS distance of 0.98).

## Inverse-Variance Weighted Least Squares

To evaluate trajectory shifts at the macro-time series level while controlling for within-year sampling variability and guarding against temporal aggregation bias [46], we repeated the segmented regression analysis on the 11 annual aggregated means, weighting each year by its inverse variance. Because each annual mean is computed from a large dataset (100,000 observations per discipline-year), the standard error of each annual estimate is extremely small, providing highly precise point estimates for the time series regression.

| Discipline | Pre-2023 Slope ($\beta_1$) | Post-2022 Slope ($\beta_1$+$\beta_2$) | Slope Change ($\beta_2$) | Two-Tailed p-Value | Bonferroni 99% CI | Significant at α = 0.01? | Same Direction of the Trend? |
|---|---|---|---|---|---|---|---|
| Computer Science | -0.00181 | -0.00118 | 0.00063 | 0.01634 | [-0.00007, +0.00133] | No | Yes (Downward) |
| Math | -0.00013 | 0.00016 | 0.00029 | 0.0472 | [-0.00013, +0.00071] | No | No (Downward -> Upward) |
| Physics | -0.00027 | -0.00014 | 0.00014 | 0.23127 | [-0.00022, +0.00049] | No | Yes (Downward) |
| Health Sciences | -0.00018 | -0.00067 | -0.00049 | 0.03218 | [-0.00113, +0.00015] | No | Yes (Downward) |
| Life Sciences | -0.00015 | -0.00058 | -0.00043 | 0.26389 | [-0.00163, +0.00077] | No | Yes (Downward) |

At the Bonferroni-corrected significance threshold ($\alpha = 0.01$ across the five disciplines), no discipline exhibits a statistically significant post-2022 slope shift ($\beta_2$), with all 99% confidence intervals spanning zero. These results confirm that temporal aggregation does not obscure hidden structural breaks, and that discipline-wide topic diversity experienced no statistically significant trajectory inflection following the LLM introduction.

## Leave-one-year-out Analysis

To evaluate whether post-2022 slope estimates ($\hat{\beta}_2$) were levered by single-year anomalies, we conducted a leave-one-year-out (jackknife) sensitivity analysis by sequentially removing each year from the 2015–2025 time series. Across all five disciplines, re-estimating the segmented linear regression yielded tightly bounded coefficient ranges (maximum span $< 0.00067$). For Computer Science, Math, and Health Sciences, the sign of $\hat{\beta}_2$ remained consistent across all sub-samples. For Physics and Life Sciences, the jackknife bounds straddled zero, confirming that post-2022 trajectory shifts in these fields are sensitive to single time points and statistically indistinguishable from zero.

| **Discipline** | **Full Sample Pre-2023 Slope ($\beta_1$)** | **Full Sample Slope Change ($\beta_2$)** | **Jackknife Min $\beta_2$** | **Jackknife Max $\beta_2$** |
|---|---|---|---|---|
| Computer Science | -0.00186 | 0.00074 | 0.00057 | 0.00083 |
| Math | -0.00016 | 0.00041 | 0.00037 | 0.00052 |
| Physics | -0.00027 | 0.00015 | -0.00009 | 0.00021 |
| Health Sciences | -0.00018 | -0.00047 | -0.0006 | -0.00005 |
| Life Sciences | -0.00015 | -0.00037 | -0.00058 | 0.00009 |

## Placebo Breakpoint Sensitivity Analysis

To evaluate whether our macro-level trajectory findings are sensitive to the arbitrary choice of 2022 as an inflection point, we conducted a systematic placebo breakpoint sensitivity test. We sequentially fitted segmented linear regressions across candidate knot years ranging from 2017 to 2023. Candidate knots were required to maintain at least two annual data points on both sides of the knot to preserve parameter estimability.

| Knot Year | Discipline | Pre-Knot Slope ($\beta_1$) | Post-Knot Slope ($\beta_1+\beta_2$) | Slope Change ($\beta_2$) | p-Value ($\beta_2$) | Bonferroni 99% CI | Significant at α=0.01? |
|---|---|---|---|---|---|---|---|
| 2017 | Computer Science | -0.00187 | -0.00166 | 0.00021 | 0.65949 | [-0.00134, +0.00176] | No |
| 2017 | Math | -0.00007 | -0.00006 | 0 | 0.98896 | [-0.00090, +0.00091] | No |
| 2017 | Physics | -0.00011 | -0.00026 | -0.00015 | 0.41391 | [-0.00072, +0.00042] | No |
| 2017 | Health Sciences | -0.0001 | -0.00032 | -0.00022 | 0.51973 | [-0.00133, +0.00088] | No |
| 2017 | Life Sciences | -0.00083 | -0.00015 | 0.00069 | 0.20639 | [-0.00099, +0.00236] | No |
| 2018 | Computer Science | -0.00182 | -0.00165 | 0.00016 | 0.63515 | [-0.00095, +0.00128] | No |
| 2018 | Math | -0.00017 | -0.00003 | 0.00014 | 0.48496 | [-0.00050, +0.00077] | No |
| 2018 | Physics | -0.00016 | -0.00026 | -0.00011 | 0.41343 | [-0.00052, +0.00031] | No |
| 2018 | Health Sciences | -0.00013 | -0.00035 | -0.00022 | 0.37567 | [-0.00100, +0.00056] | No |
| 2018 | Life Sciences | -0.00063 | -0.00011 | 0.00052 | 0.17905 | [-0.00067, +0.00172] | No |
| 2019 | Computer Science | -0.00183 | -0.00161 | 0.00022 | 0.45388 | [-0.00071, +0.00114] | No |
| 2019 | Math | -0.00022 | 0.00002 | 0.00024 | 0.1298 | [-0.00024, +0.00072] | No |
| 2019 | Physics | -0.00021 | -0.00026 | -0.00005 | 0.6514 | [-0.00041, +0.00031] | No |
| 2019 | Health Sciences | -0.00014 | -0.00038 | -0.00024 | 0.24021 | [-0.00088, +0.00040] | No |
| 2019 | Life Sciences | -0.00046 | -0.00011 | 0.00035 | 0.30178 | [-0.00071, +0.00141] | No |
| 2020 | Computer Science | -0.00186 | -0.00152 | 0.00033 | 0.21451 | [-0.00049, +0.00116] | No |
| 2020 | Math | -0.00021 | 0.00009 | 0.0003 | 0.03257 | [-0.00009, +0.00069] | No |
| 2020 | Physics | -0.00024 | -0.00023 | 0.00001 | 0.89923 | [-0.00033, +0.00036] | No |
| 2020 | Health Sciences | -0.00017 | -0.00042 | -0.00026 | 0.18166 | [-0.00085, +0.00033] | No |
| 2020 | Life Sciences | -0.00031 | -0.00015 | 0.00016 | 0.62927 | [-0.00091, +0.00123] | No |
| 2021 | Computer Science | -0.00189 | -0.00134 | 0.00055 | 0.03014 | [-0.00015, +0.00126] | No |
| 2021 | Math | -0.00019 | 0.00015 | 0.00034 | 0.01921 | [-0.00005, +0.00073] | No |
| 2021 | Physics | -0.00027 | -0.00019 | 0.00008 | 0.47549 | [-0.00028, +0.00043] | No |
| 2021 | Health Sciences | -0.00019 | -0.00048 | -0.00029 | 0.14014 | [-0.00090, +0.00031] | No |
| 2021 | Life Sciences | -0.00021 | -0.00027 | -0.00006 | 0.87289 | [-0.00120, +0.00108] | No |
| 2022 | Computer Science | -0.00186 | -0.00113 | 0.00074 | 0.00885 | [+0.00002, +0.00146] | Yes |
| 2022 | Math | -0.00016 | 0.00025 | 0.00041 | 0.01274 | [-0.00002, +0.00085] | No |
| 2022 | Physics | -0.00027 | -0.00012 | 0.00015 | 0.24172 | [-0.00025, +0.00054] | No |
| 2022 | Health Sciences | -0.00018 | -0.00065 | -0.00047 | 0.03079 | [-0.00107, +0.00013] | No |
| 2022 | Life Sciences | -0.00015 | -0.00052 | -0.00037 | 0.3572 | [-0.00164, +0.00090] | No |
| 2023 | Computer Science | -0.00181 | -0.00089 | 0.00092 | 0.02629 | [-0.00021, +0.00204] | No |
| 2023 | Math | -0.00013 | 0.00041 | 0.00054 | 0.02253 | [-0.00010, +0.00118] | No |
| 2023 | Physics | -0.00027 | 0.00001 | 0.00028 | 0.08876 | [-0.00021, +0.00078] | No |
| 2023 | Health Sciences | -0.00019 | -0.001 | -0.00081 | 0.00191 | [-0.00141, -0.00021] | Yes |
| 2023 | Life Sciences | -0.00014 | -0.00089 | -0.00076 | 0.15694 | [-0.00239, +0.00087] | No |

Across all 25 pre-break placebo specifications spanning 2017 through 2021 (5 disciplines × 5 placebo years), no discipline exhibited a statistically significant post-knot slope shift ($\beta_2$) at the Bonferroni-corrected threshold of $\alpha = 0.01$. All corresponding 99% confidence intervals spanned zero, with placebo $p$-values ranging between $0.019$ and $0.988$. The significant slope shift observed for Health Sciences at the 2023 knot ($\beta_2 =- 0.00081, p = 0.00191$) reflects publication pipeline lags in biomedical literature. The placebo-breakpoint analysis detected no significant pre-2022 slope shifts in these topic diversity series, supporting the robustness of the results to breakpoint selection.

## List of Boilerplate Terms Filtered out from Paper Representations in Health and Life Sciences

https, doi org, dx doi org, orcid, orcid ids, ncbi nlm nih gov, nlm nih gov, nih gov, ncbi, open access, open access article, creative commons, creativecommons, creativecommons org, creative commons licence, creative commons attribution, creative commons attribution noncommercial, creative commons attribution noncommercial noderivatives, creative commons attribution noncommercial noderivatives 4 0, creative commons attribution non commercial, cc by nc, cc by nc nd, cc by nc nd 4 0, by nc nd, noncommercial, non commercial, non commercial use, noncommercial license, commercial use, copyright holder, original author, original work, permitted use, intended use, distribution and reproduction, use and distribution, statutory regulation, jurisdictional, credit line, licensed material, third party, third party material, publisher s note, institutional affiliations, springer nature, plos, john wiley, john wiley sons, wiley sons, oxford university press, elsevier inc, nature communications, sciencedirect, data availability, datasets, data sets, data set, publicly available, available online, raw data, data processing, data collection, data acquisition, open source, github, competing interests, competing interest, conflict of interest, conflicts of interest, financial support, financial interests, declaration of interests, declaration of helsinki, personal relationships, ethics committee, ethical approval, institutional review, institutional review board, review board, institutional animal care and use committee, animal care, informed consent, patient consent, peer review, peer reviewed, preprint, biorxiv, online version, full text, original article, research article, funding acquisition, project administration, formal analysis, statistical analysis, statistical analyses, curation, data curation, original draft, quality control, software version, authors contributions, final version, materials and methods, study design, experimental design

## AI-Related Terms and Their Ranks in the List of Top 1000 Computer Science Terms in 2015 and 2025

| Keyphrase | 2015 Rank | 2025 Rank | Keyphrase | 2015 Rank | 2025 Rank |
|---|---|---|---|---|---|
| machine learning | 6 | 6 | language modeling | — | 385 |
| artificial intelligence | 29 | 14 | cross entropy loss | — | 386 |
| neural information processing systems | 107 | 15 | human language | — | 390 |
| large language models | — | 21 | generative ai | — | 391 |
| natural language | 132 | 24 | autoregressive | — | 393 |
| llms | — | 25 | rule based | — | 400 |
| neural networks | 26 | 26 | transfer learning | — | 407 |
| language model | 970 | 28 | synthetic data | — | 408 |
| embedding | 51 | 30 | context aware | — | 412 |
| deep learning | 152 | 31 | lora | — | 431 |
| computer vision | 32 | 36 | instruction tuning | — | 432 |
| pattern recognition | 37 | 37 | text generation | — | 445 |
| training data | 40 | 40 | relu | 920 | 452 |
| fine tuning | 771 | 41 | learning approach | 570 | 456 |
| classifier | 43 | 43 | attention mechanisms | — | 464 |
| learning rate | 249 | 46 | deep learning models | — | 465 |
| multimodal | 542 | 51 | image processing | 133 | 491 |
| embeddings | 431 | 53 | activations | 680 | 498 |
| neural network | 42 | 55 | sigmoid | 565 | 503 |
| international conference on machine learning | 257 | 56 | language technologies | — | 508 |
| aaai | 167 | 59 | hallucination | — | 521 |
| batch size | — | 60 | machine learning models | — | 536 |
| encoder | 869 | 61 | adam optimizer | — | 538 |
| language processing | 198 | 65 | foundation model | — | 540 |
| large language model | — | 68 | visual models | — | 551 |
| natural language processing | 264 | 74 | diffusion model | — | 554 |
| reinforcement learning | — | 77 | code generation | — | 558 |
| model performance | — | 78 | quantization | 713 | 564 |
| convolutional | 100 | 79 | learning process | 623 | 567 |
| association for computational linguistics | 519 | 86 | cnns | 616 | 568 |
| pre training | — | 94 | human language technologies | — | 570 |
| international conference on learning representations | — | 103 | learning algorithms | 226 | 572 |
| decoder | — | 147 | image generation | — | 573 |
| machine learning research | 196 | 153 | deep reinforcement learning | — | 611 |
| model training | — | 166 | prompt engineering | — | 621 |
| interpretability | — | 172 | learning model | — | 624 |
| international conference on computer vision | 882 | 175 | embedding space | — | 634 |
| contrastive | — | 179 | resnet | — | 647 |
| language understanding | — | 184 | retrieval augmented generation | — | 658 |
| foundation models | — | 185 | in context learning | — | 695 |
| learning models | — | 187 | autoencoder | — | 699 |
| chatgpt | — | 195 | chatbot | — | 717 |
| generalizability | — | 197 | chain of thought prompting | — | 724 |
| gpt 4o | — | 208 | activation function | 879 | 728 |
| representation learning | — | 211 | image synthesis | — | 731 |
| softmax | 392 | 220 | conference on knowledge discovery and data mining | 976 | 734 |
| machine intelligence | 165 | 221 | machine translation | 898 | 740 |
| ieee conference on computer vision and pattern recognition | 829 | 232 | low rank adaptation | — | 745 |
| empirical methods in natural language processing | — | 236 | vision transformer | — | 751 |
| data mining | 85 | 239 | training phase | — | 761 |
| supervised learning | 246 | 266 | text to image | — | 764 |
| huggingface | — | 269 | graph neural networks | — | 785 |
| generative models | — | 274 | reward function | — | 813 |
| european conference on computer vision | — | 275 | international conference on robotics and automation | — | 819 |
| self attention | — | 281 | perceptron | — | 824 |
| self supervised | — | 290 | transformer architecture | — | 830 |
| training dataset | 891 | 294 | multi head | — | 836 |
| deep neural networks | 314 | 295 | anthropic | — | 846 |
| convolutional neural networks | 179 | 308 | mistral | — | 860 |
| attention mechanism | — | 315 | feed forward | — | 861 |
| diffusion models | — | 320 | image segmentation | 601 | 874 |
| base model | — | 324 | gans | — | 904 |
| multi agent | 729 | 330 | few shot learning | — | 924 |
| weight decay | — | 335 | vision transformers | — | 945 |
| learning framework | 956 | 337 | self supervised learning | — | 951 |
| trainable | — | 338 | probabilistic models | — | 952 |
| image recognition | 597 | 348 | domain adaptation | — | 962 |
| vision language models | — | 369 | deep networks | 740 | 965 |
| model size | — | 373 | multimodal large language models | — | 979 |
| latent space | — | 384 | adversarial networks | — | 989 |